\documentclass[a4paper, twocolumn,10pt]{asme2ej}

\usepackage{epsfig} %% for loading postscript figures

\usepackage{graphicx}

\usepackage{amssymb}

\usepackage{amsmath}
\newcommand{\vect}[1]{\left\{\begin{matrix}#1\end{matrix}\right\}}
\newcommand{\vq}{{\mathbf{q}}}
\newcommand{\vdq}{{\dot{\mathbf{q}}}}
\newcommand{\vddq}{{\ddot{\mathbf{q}}}}
\newcommand{\vtau}{{\mathbf{\tau}}}
\newcommand{\vphi}{{\mathbf{\phi}}}
\newcommand{\vchi}{{\mathbf{\chi}}}
\newcommand{\etal}{\textit{et al.}}

\newcommand{\vOmega}{{\boldsymbol{\Omega}}}

\usepackage[dvips]{color}

\title{Drawbacks and alternatives to the numerical calculation of the base inertial parameters expressions for low mobility mechanisms}
\author{Xabier Iriarte\thanks{Address all correspondence to this author.}
    \affiliation{
	Departamento de Ingenier\'ia\\
	Mec\'anica, Energ\'etica y de Materiales.\\
	Universidad P\'ublica de Navarra,\\ 31006 Pamplona, Navarra, Spain. \\
    Email: xabier.iriarte@unavarra.es
    }	
}

\author{Javier Ros
    \affiliation{
	Departamento de Ingenier\'ia\\
	Mec\'anica, Energ\'etica y de Materiales.\\
	Universidad P\'ublica de Navarra,\\ 31006 Pamplona, Navarra, Spain. \\
    Email: jros@unavarra.es
    }	
}

\author{Aitor Plaza
    \affiliation{
	Departamento de Ingenier\'ia\\
	Mec\'anica, Energ\'etica y de Materiales.\\
	Universidad P\'ublica de Navarra,\\ 31006 Pamplona, Navarra, Spain. \\
    Email: aitor.plaza@unavarra.es
    }	
}

\author{Jokin Aginaga
    \affiliation{
	Departamento de Ingenier\'ia\\
	Mec\'anica, Energ\'etica y de Materiales.\\
	Universidad P\'ublica de Navarra,\\ 31006 Pamplona, Navarra, Spain. \\
    Email: jokin.aginaga@unavarra.es
    }	
}

\author{Vicente Mata
    \affiliation{
	Centro de Investigaci\'on de\\
	Tecnolog\'ia de Veh\'iculos.\\
	Universidad Polit\'ecnica de Valencia,\\
	C/Camino de Vera s/n,\\
	46022 Valencia, Spain \\
    Email: vmata@mcm.upv.es
    }	
}

\begin{document}

\maketitle    

%%%%%%%%%%%%%%%%%%%%%%%%%%%%%%%%%%%%%%%%%%%%%%%%%%%%%%%%%%%%%%%%%%%%%%
\begin{abstract}
{\it Base inertial parameters constitute a minimal inertial parametrization of mechanical systems that is of interest, for example, in parameter estimation and model reduction. Numerical and symbolic methods are available to determine their expressions. In this paper the problems associated with the numerical determination of the base inertial parameters expressions in the context of low mobility mechanisms are analyzed and discussed through and example. To circumvent these problems two alternatives are proposed: a variable precision arithmetic implementation of the customary numerical algorithm and the application of a general symbolic method. Finally, the advantages of both approaches compared to the numerical one are discussed in the context of the proposed low mobility example.
}
\end{abstract}

%%%%%%%%%%%%%%%%%%%%%%%%%%%%%%%%%%%%%%%%%%%%%%%%%%%%%%%%%%%%%%%%%%%%%%
% \begin{nomenclature}
% \entry{A}{You may include nomenclature here.}
% \entry{$\alpha$}{There are two arguments for each entry of the nomemclature environment, the symbol and the definition.}
% \end{nomenclature}
% 
% The primary text heading is  boldface and flushed left with the left margin.  The spacing between the  text and the heading is two line spaces.

%%%%%%%%%%%%%%%%%%%%%%%%%%%%%%%%%%%%%%%%%%%%%%%%%%%%%%%%%%%%%%%%%%%%%%

% ------------------------------------------------------------------------------------
% ------------------------------------------------------------------------------------
	\section{Introduction}
% ------------------------------------------------------------------------------------
% ------------------------------------------------------------------------------------

Realistic simulation and optimization of mechanical systems require accurate and reliable dynamic models. Additionally, advanced mechatronic systems usually rely on model-based control strategies, and their performance critically depends on the accuracy of the models. In recent decades, great efforts have been put into modeling mechanical systems, using advanced and efficient numerical formulations \cite{Featherstone2008,GarciaJalon1994,Samin2004}, complex friction and damping models \cite{Awrejcewicz2005}, considering flexibility \cite{Bauchau2011}, etc. However, when representative numerical values of some parameters are not known with the required level of accuracy \cite{Atkeson1986}, the resulting models tend to have limited predictive ability and increasing their complexity does not improve their performance.

While geometrical parameters are usually known with enough precision, information on inertial and frictional parameters is difficult to obtain by a simple inspection of the system. If accurate prediction models are needed, they will have to be calibrated in order to accurately estimate the output of the system. This calibration can be usually accomplished by model identification~\cite{Ljung1998} or parameter estimation ~\cite{vandenBos2007} techniques, in which the inputs and outputs of the system are measured at different experimental points and, based on these, the parameters of the model are estimated. When the output of the model can be written as a linear combination of the unknown parameters, estimates of these parameters can, a priori, be calculated in a straightforward manner \cite{Khalil2002}.

For mechanical systems, if the inertia tensors are defined in non-centroidal reference frames and the center of gravity positions are expressed in terms of the first moments of inertia of each body, the dynamic equations of a mechanical system can be written in linear form with respect to the inertial parameters~\cite{Shome1998,Maes1989}:
\begin{equation}	\label{eq::dyn_equations}
	K(\vq,\vdq,\vddq)\vphi=\vtau
\end{equation}
where $\vq$, $\vdq$ and $\vddq$ are the generalized coordinates, velocities and accelerations and $\vtau$ is the vector of external generalized forces applied to the system. The vector of inertial parameters, $\vphi$, contains the mass, first moments, and second moments and products of inertia of each body, as shown in Eqn.~(\ref{eq::inertial_parameters_ith_body}) for the $i^{th}$ body.
\begin{equation}	\label{eq::inertial_parameters_ith_body}
	\vphi_i=(m_i, mx_i, my_i, mz_i, Ixx_i, Ixy_i, Ixz_i, Iyy_i, Iyz_i, Izz_i)^T.
\end{equation}

For systems with constraints, the open-loop dynamic equations have to be multiplied by an orthogonal complement of the constraint Jacobian in order to obtain an equation with the same structure as Eqn.~(\ref{eq::dyn_equations}) but with as many rows as the number of degrees of freedom of the mechanism.

The parameter estimation equations for an open- or closed-loop mechanism can be obtained by assembling the system of Eqn.~(\ref{eq::dyn_equations})  for a set of $N$ different experimental points:
\begin{equation}	\label{eq::system_equation}
	\begin{bmatrix}
		K(\vq_1,\vdq_1,\vddq_1)	\\
		K(\vq_2,\vdq_2,\vddq_2)	\\
		\vdots				\\
		K(\vq_N,\vdq_N,\vddq_N)	\\
	  \end{bmatrix}\vphi=W\vphi=\vchi=
	\vect{\vtau_1 \\ \vtau_2 \\ \vdots \\ \vtau_N}.
\end{equation}
where $W$ is the so-called observation matrix. If the models for stiffness, damping, friction and other types of constitutive forces are linear with respect to their respective parameters, these parameters can also be included in the parameter vector $\vphi$ in Eqn.~(\ref{eq::system_equation}) ~\cite{Grotjahn2001}.

The observation matrix $W$ is usually found to be rank deficient. This rank deficiency is related to the limitation of movement of bodies that, in turn, is associated with the presence of joints. Consequently, it is impossible to find a least squares solution for the system of Eqn.~(\ref{eq::system_equation}). Nevertheless, physically meaningful solutions can be obtained for sets of linear combinations of the parameters in $\vphi$. A possible choice of such estimable sets of linear combinations of parameters are the so-called \emph{base inertial parameters}~\cite{Mayeda1990}.
These parameters are written in terms of the inertial parameters as:
\begin{equation}
      \vphi_b=\vphi_1+\beta \vphi_2
\end{equation}
where $\vphi^T=(\vphi_1^T,\vphi_2^T)$ is a splitting of the inertial parameter vector and $\beta$ is a constant matrix dependent on the geometrical parameters of the mechanism. Determining the expressions of the base inertial parameters is equivalent to determine the matrix $\beta$. In this paper we will refer to calculating $\beta$ as to determining the expressions for the base inertial parameters.

The methods used to calculate expressions for the base inertial parameters can be classified into numerical and symbolic. The numerical method of Gautier~\cite{Gautier1991} to determine these expressions has shown to be easy to apply and general, but its application is limited by the machine precision. Methods to optimize the choice of experimental points used to assemble $W$ improving its numerical conditioning  have been proposed~\cite{DiazRodriguez2009,Park2006,Swevers1996,Swevers1997}. Despite this, the determination of base inertial parameters expressions remains a problematic issue for mechanisms with low mobility.

On the other hand, symbolic methods have been proposed to determine the barycentric~\cite{Fisette1996}, minimum~\cite{Gautier1990a}, and base~\cite{Chen2002a,Khalil1995} inertial parameters. These methods provide the symbolic expressions for the base inertial parameters in terms of the geometric properties of the model, avoiding potential drawbacks of numerical approaches.
A noteworthy work in this field is that of Ros \etal~\cite{Ros2012}, the procedure proving to be simple, intuitive and general.

In the algorithm of Gautier~\cite{Gautier1991}, the rank of matrix $W$ is calculated numerically as the number of singular values of $W$ that are larger than a certain machine dependent tolerance. For systems with low mobility, when using double-precision (DP) arithmetic there is frequently no natural cut-off value that clearly separates the singular values larger and smaller than the given tolerance. In this situation the calculated rank depends on truncation errors and therefore Gautier's procedure can yield incorrect results.
In order to overcome this problem, a variable-precision (VP) arithmetic implementation of Gautiers algorithm is proposed and presented in this paper together with the symbolic method of Ros \etal~\cite{Ros2012}. Along with the numerical method of Gautier, the VP base method and the symbolic one have been applied to a SLA suspension, and the obtained base inertial parameters expressions have been compared and discussed.

In Section~\ref{sec::gautier}, the numerical algorithm of Gautier is briefly described.
Its standard and proposed VP arithmetic implementations are explained in Section~\ref{sec::gautier_implementation}.
Afterwards, in Section~\ref{sec::symbolic_method}, the symbolic procedure of Ros \etal~\cite{Ros2012} is briefly reviewed.
After introducing the three methods to determine the base inertial parameters expressions, the SLA suspension under study is described in Section~\ref{sec::the_system}.
In Section~\ref{sec::results} the base inertial parameters expressions are calculated in detail with the three methods.
In Section~\ref{sec::discussion} the inability of the numerical procedure of Gautier to correctly obtain the expressions of the base inertial parameters when using DP arithmetic is illustrated. Moreover, it is shown that the VP implementation of Gautiers algorithm correctly obtains the values of the expressions of the base inertial parameters.
These results are in turn used to ascertain the correctness of the proposed VP implementation and the symbolic procedure, and the benefits and drawbacks of symbolic vs. numerical methods are discussed. Finally, some conclusions are drawn.

% -----------------------------------------------------------------------------------------------------------------
% -----------------------------------------------------------------------------------------------------------------
	\section{Gautier's numerical procedure}	\label{sec::gautier}
% -----------------------------------------------------------------------------------------------------------------
% -----------------------------------------------------------------------------------------------------------------

In the following we briefly review the numerical procedure of Gautier ~\cite{Gautier1991} to determine the expressions of the base inertial parameters.

First, note that Eqn.~(\ref{eq::system_equation}) are obtained out from system (\ref{eq::dyn_equations}) writing it for a set of $N$ experimental points.
Gathering those experimental points in an exciting trajectory~\cite{Swevers1997}, matrix $W$ can be evaluated and further decomposed into $W=U \, S \, V^T$ using the singular value decomposition.
Right-multiplying $W$ by $V$:
\begin{equation}	\label{eq::svd_modif}
	W \begin{bmatrix}
		V_1	&	V_2
	  \end{bmatrix}
	= U \begin{bmatrix}
		\Sigma	&	0	\\
		0	&	0	
	    \end{bmatrix}
\end{equation}
is obtained, where $\Sigma$ is full rank diagonal.  Since $WV_2=0$, the following equation will hold for any vector $\vphi_a$ :
\begin{equation}	\label{eq::modelo_reducido}
	W\vphi=W(\vphi+V_2 \vphi_a)=W \vphi_R
\end{equation}
Vector $(\vphi+V_2 \vphi_a)$ is reordered as $\Pi^T (\vphi+V_2 \vphi_a) = [\vphi_1^T,\vphi_2^T]^T$, where $\Pi$ is a permutation matrix such that in
\begin{equation}
	\Pi^T V_2=
	\begin{bmatrix}
		V_{21}	\\
		V_{22}
	\end{bmatrix}
\end{equation}
$V_{22}$ is square and full rank.
Therefore $\vphi_a$ can be chosen to make the last $dim(\vphi)-rank(W)$ elements of $\vphi_2$  zero, making $(\vphi+V_2 \vphi_a)=[\vphi_{b}^T,\mathbf{0}]^T$ and
leading to the base parameters expressions:
\begin{equation}	\label{eq::param_base_definition}
	\vphi_b=\vphi_1-V_{21} V_{22}^{-1} \vphi_2 = \vphi_1+\beta \vphi_2.
\end{equation}
Note that
\begin{equation}	\label{eq::reduced_model_W}
	W \, \Pi \,\Pi^T \, \vphi = W \, \Pi \, [\vphi_b^T, \mathbf{0}]^T= W_b \, \vphi_b=\vchi.
\end{equation}

% -----------------------------------------------------------------------------------------------------------------
% -----------------------------------------------------------------------------------------------------------------
	\section{Gautier's algorithm implementation using DP and VP arithmetic}	\label{sec::gautier_implementation}
% -----------------------------------------------------------------------------------------------------------------
% -----------------------------------------------------------------------------------------------------------------

The DP implementation of the algorithm of Gautier is straightforward. All the calculations are done with customary double precision arithmetic and the results are obtained. For the VP implementation, care must be taken to assure that every single calculation is performed with VP arithmetic. This can be challenging and depends on the underlying system used. Authors have performed two different implementations on two different algebra systems to validate the correctness of the approach. Calculation of the points of the trajectory, matrix assemble, SVD calculation (which takes most of the computational time) and matrix manipulation have all to be done using VP arithmetic software in order to ensure that the solution is calculated with the desired number of significant digits.

In the presence of constraints, the determination of the generalized coordinates out of the independent coordinates set using the Newton-Raphson algorithm has to be repeated until the dependent generalized coordinates have been calculate correctly with the desired digits. An important aspect to consider is that the calculation of the singular values of matrix $W$ has to be repeated several times, with an increasing number of digits, until the ridge of the singular values is evident. At that point, the singular values after the ridge converge to zero as the number of digits increases, while the singular values before the ridge keep constant.

% -----------------------------------------------------------------------------------------------------------------
% -----------------------------------------------------------------------------------------------------------------
	\section{Symbolic Base Inertial Parameter Calculation Method}	\label{sec::symbolic_method}
% -----------------------------------------------------------------------------------------------------------------
% -----------------------------------------------------------------------------------------------------------------

As an alternative method to Gautiers numerical procedure, symbolic methods can also provide the expressions of the base inertial parameters. However, they do not need to use an exciting trajectory and the results are obtained in terms of the geometrical parameters of the mechanism.

For the sake of completeness, let us briefly review the algorithm algorithm of Ros \etal~\cite{Ros2012} to calculate the symbolic base inertial parameters expressions of a mechanism.

\subsection{Representation of the Lagrangian in terms of inertia monopoles, dipoles and quadrupoles}	
% -----------------------------------------------------------------------------------------------------------------

For systems with ideal joints in which friction is not dependent on the constraint reactions, Lagrangian mechanics says that inertial contributions to the dynamics of motion (kinetics) solely depend on the non-constant terms of the kinetic, $T$, and gravitational potential, $V$, energy of the system through the Lagrangian, $L=T-V$.

The inertial contributions to the Lagrangian of a rigid body $Body_i$, can be expressed as:

\begin{equation}
 L(Body_i)=\frac{1}{2}  \mathbf{V}_{G_i} \cdot (m_i \mathbf{V}_{G_i} ) + \frac{1}{2} {\vOmega}_i  \cdot ( \mathbf{I}_{G_i}\vOmega_i ) -(-m_i \overline{OG_i} \cdot \mathbf{g}),
\end{equation}
where $\mathbf{g}$ is the gravity vector, $\mathbf{V}_{G_i}$  and $\vOmega_i$ are the velocity and angular velocity vectors of the center of inertia and body respectively, and $m_i$ and $\mathbf{I}_{G_i}$ are the mass and inertia tensor of the body.

If $B_i$ is a point of $Body_i$, the position and velocity of the center of inertia can be expressed as:

\begin{eqnarray}
 \overline{OG_i}	&=&\overline{OB_i}+\overline{B_i G_i} \\
 \mathbf{V}_{G_i} &=& \mathbf{V}_{B_i} + \widetilde{\vOmega}_i \overline{B_iG_i},
\end{eqnarray}
where $\widetilde{\vOmega}_i$ represents the skew symmetric matrix associated with $\vOmega_i$. Then the Lagrangian can be rewritten as:

\begin{eqnarray}
 L(Body_i)&=&\left[\frac{1}{2}  (\mathbf{V}_{B_i} \cdot \mathbf{V}_{B_i})+\overline{OB_i} \cdot \mathbf{g}\right] m_i\\
		 &+&\left[ - \widetilde{\vOmega}_i  \mathbf{V}_{B_i} + \mathbf{g} \right] \cdot m_i \overline{B_iG_i}
 +  \frac{1}{2} \vOmega_i \cdot ( \mathbf{I}_{B_i} \vOmega_i )\nonumber
\end{eqnarray}
where $\mathbf{I}_{B_i}=\mathbf{I}_{G_i}-m_i \widetilde{B_iG_i} \widetilde{B_iG_i}$.

\begin{figure}[htb]
	\begin{center}
		\includegraphics[width=0.45\textwidth]{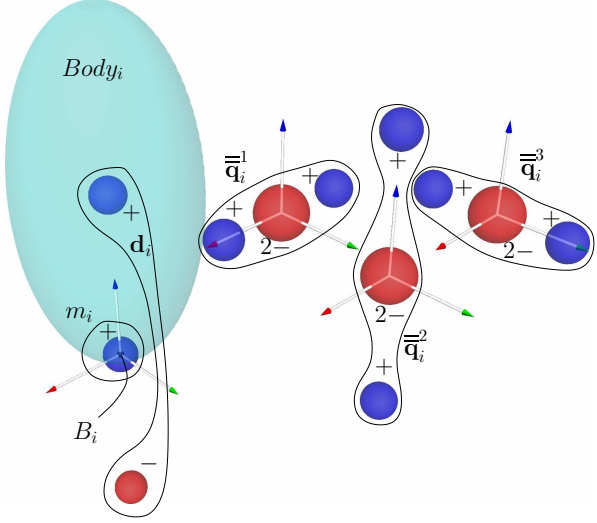}
		\caption{Ellipsoid of revolution and its multipole representation with respect to $B_i$. Spheres represent point masses, the color their sign (blue positive, red negative) and the size their modulus. Quadrupole position is not relevant.} \label{fig:mdq}
	\end{center}
\end{figure}

Now the notion of multipole is geometrically introduced by means of Fig.~\ref{fig:mdq} in which a body, $Body_i$ (in the example, an  ellipsoid of revolution ), and its inertial properties are described in terms of multipoles at $B_i$:
\begin{itemize}
	\item[-] $m_i$ is the \emph{monopole} at $B_i$: it is defined as a positive point mass at $B_i$.
	\item[-] $\mathbf{d}_i$ is the \emph{dipole} at $B_i$: it is defined as a pair of point masses, one positive and the other negative, with the same modulus and equidistant from $B_i$, such that $\mathbf{d}_i=m_i \overline{B_iG_i}$.
	\item[-] $\overline{\overline{\mathbf{q}}}_i^1$, $\overline{\overline{\mathbf{q}}}_i^2$, and $\overline{\overline{\mathbf{q}}}_i^3$ are three elemental \emph{quadrupoles} along independent directions: each of them is composed of two dipoles of equal magnitude and opposite sense, so that $\mathbf{I}_{B_i}=\overline{\overline{\mathbf{q}}}_i^1+\overline{\overline{\mathbf{q}}}_i^2+\overline{\overline{\mathbf{q}}}_i^3$. The quadrupoles are intentionally drawn in different positions, to emphasize the irrelevance of quadrupole position.
\end{itemize}

These three multipoles are independent of each other since changing the value of one of them does not alter the others. This independence will be a key feature when calculating the base parameters expressions.

With these definitions of multipoles, the Lagrangian can be rewritten as the sum of the contributions of three different types of multipoles:
\begin{eqnarray}
 L(Body_i)&=&\left[\frac{1}{2}  (\mathbf{V}_{B_i} \cdot \mathbf{V}_{B_i})+\overline{OB_i} \cdot \mathbf{g}\right] m_i\\
			&+&\left[ - \widetilde{\vOmega}_i  \mathbf{V}_{B_i} + \mathbf{g} \right] \cdot \mathbf{d}_i
 +  \frac{1}{2} \vOmega_i \cdot ( \sum_k \overline{\overline{\mathbf{q}}}_i^k  \vOmega_i ).\nonumber
\label{eq:kin_ener_rewriten}
\end{eqnarray}

\subsection{Inertial properties without effect on the kinetics}	 \label{sec:non_affecting_inertial_properties}
% -----------------------------------------------------------------------------------------------------------------

In general not all the inertial properties of the different bodies of a system will have an effect on the dynamics of motion of a rigid-body mechanical system. That is, some of them will not appear in the equations of motion.

From Eqn.~(\ref{eq:kin_ener_rewriten}) it is easy to see which are the necessary conditions for a given multipole of a generic body $Body_i$ to disappear from the equations of motion of the system:
 \begin{enumerate}
	\item Monopole $m_i$ disappears, when $\mathbf{V}_{B_i}=\mathbf{0}$ and $\overline{OB_i} \cdot \mathbf{g}=cst$.
	\item Dipole $\mathbf{d}_i$ disappears, in the direction of $\vOmega_i$, when $\mathbf{g}\cdot\mathbf{d}_i=cst$.
	\item Dipole $\mathbf{d}_i$ disappears, in the direction of $\mathbf{V}_{B_i}$, when $\mathbf{g}\cdot\mathbf{d}_i=cst$.
	\item Dipole $\mathbf{d}_i$ disappears, in any direction, if $\vOmega_i=\mathbf{0}$ or $\mathbf{V}_{B_i}=\mathbf{0}$, when $\mathbf{g}\cdot\mathbf{d}_i=cst$.
	\item All the products of inertia and moments of inertia $\overline{\overline{\mathbf{q}}}_i$ disappear in the directions orthogonal to $\vOmega_i$
\footnote{This condition is more easily expressed in terms of inertia tensor components. Note that each of the components of the inertia tensor can be related to quadrupoles.}.
	\item Quadrupole $\overline{\overline{\mathbf{q}}}_i$ disappears, in any direction if $\vOmega_i=\mathbf{0}$.
 \end{enumerate}

For a parameter to be removed, the multipoles should be constant with respect to the points $B_i$ and $B_j$ and in the respective orientation in which they are defined.

\subsection{Inertia transfers with no effect on the kinetics}	\label{sec:inertia_transfers}
% -----------------------------------------------------------------------------------------------------------------

The kinetics or dynamics of motion of a virtual system obtained from the original one by virtually changing the mass distribution is not altered as long as the Lagrangian of the virtual system remains the same. This change in the mass distribution can be geometrically interpreted in terms of changes of the previously defined monopoles, dipoles and quadrupoles.

At a theoretical level, one of the advantages of the multipole-based representation of the Lagrangian, is that it is possible to alter the mass distribution of a body changing a given multipole, while the remaining multipoles are unaltered. Therefore, the contribution of inertial properties to the Lagrangian defined in terms of the introduced monopoles, dipoles and quadrupoles can be considered to be independent.

Due to this independence, it is easy to identify the conditions under which the inertial properties change without affecting their Lagrangian. To that end, it suffices to analyze the change of the Lagrangian of a pair of bodies. For $Body_i$ \& $Body_j$, when  a monopole $m$, a dipole $\mathbf{d}$  and a quadrupole $\overline{\overline{\mathbf{q}}}$ are transferred from $Body_j$ to $Body_i$, the change in the Lagrangian is:

\begin{eqnarray}	\label{eq::lagrangian}
      \Delta L=&+&\left(  \frac{1}{2}  (  \mathbf{V}_{B_i} \cdot \mathbf{V}_{B_i} - \mathbf{V}_{B_j} \cdot \mathbf{V}_{B_j}  ) + \overline{B_jB_i}\cdot \mathbf{g}\right)m \label{deltat} \\
      &+&( - \widetilde{\vOmega}_i \mathbf{V}_{B_i}  + \widetilde{\vOmega}_j \mathbf{V}_{B_j}   ) \cdot \mathbf{d} 
      +   \frac{1}{2} ( \vOmega_i \cdot ( \overline{\overline{\mathbf{q}}} ~\vOmega_i ) - \vOmega_j \cdot ( \overline{\overline{\mathbf{q}}} ~\vOmega_j ) ) \nonumber
\end{eqnarray}

% \begin{eqnarray}	\label{eq::lagrangian}
%       \Delta L=&+&\left(  \frac{1}{2}  (  \mathbf{V}_{B_i} \cdot \mathbf{V}_{B_i} - \mathbf{V}_{B_j} \cdot \mathbf{V}_{B_j}  ) + (\overline{OB_i}-\overline{OB_j})\cdot \mathbf{g}\right)m \label{deltat} \\
%       &+&( - \widetilde{\vOmega}_i \mathbf{V}_{B_i}  + \widetilde{\vOmega}_j \mathbf{V}_{B_j}   ) \cdot \mathbf{d} 
%       +   \frac{1}{2} ( \vOmega_i \cdot ( \overline{\overline{\mathbf{q}}} ~\vOmega_i ) - \vOmega_j \cdot ( \overline{\overline{\mathbf{q}}} ~\vOmega_j ) ) \nonumber
% \end{eqnarray}

The multipoles added to one of the bodies and removed from the other must be the same, so that the Lagrangian does not change.

In Eqn.~(\ref{eq::lagrangian}) the monopole and the dipole are added / removed at a common point, so $B_i$ and $B_j$ refer to two different points belonging to $Body_i$ and $Body_j$ respectively, but which --at least instantaneously (as in the case of contact between two bodies)-- occupy the same spatial position. By contrast, as quadrupoles have no defined positions, they can be added and removed at points with different position in space.

Obviously a given multipole transfer will not affect the dynamics of motion whenever $\Delta L=0$. According to Eqn.~(\ref{eq::lagrangian}) the multipole transfers that do not affect the kinetics are:
 \begin{enumerate}
  \item Monopole $m$ can be transferred, when $\mathbf{V}_{B_i}=\mathbf{V}_{B_j}$ and $\overline{B_iB_j} \cdot \mathbf{g}=cst$.
  \item Dipole $\mathbf{d}$ can be transferred in the direction of $\vOmega_i-\vOmega_j$, when $\mathbf{V}_{B_i}=\mathbf{V}_{B_j}$.
  \item Dipole $\mathbf{d}$ can be transferred in the direction of $\mathbf{V}_{B_i}-\mathbf{V}_{B_j}$, when $\vOmega_i=\vOmega_j$.
  \item Dipole $\mathbf{d}$ can be transferred in any direction, when $\mathbf{V}_{B_i}=\mathbf{V}_{B_j}$ and $\vOmega_i=\vOmega_j$.
  \item Quadrupole $\overline{\overline{\mathbf{q}}}$ can be transferred in the direction of $\vOmega_i-\vOmega_j$.
  \item Quadrupole $\overline{\overline{\mathbf{q}}}$ can be transferred in any direction if $\vOmega_i=\vOmega_j$.
 \end{enumerate}
For joints satisfying any of these conditions, multipoles can be transferred to cancel the corresponding multipole in the donor body while adding it to the same multipole of the acceptor. This in turn changes acceptor inertial properties that are then expressed as linear combinations of the inertial properties of original acceptor and those of the donor. Accordingly, the process reduces the total number of inertial parameters of the bodies of the transformed mechanism while it preserves the kinetics, and thus can be used to reduce the dynamical model and determine the base parameters expressions.

As with the conditions presented in Section~\ref{sec:non_affecting_inertial_properties}, the fulfillment of just the aforementioned rules or conditions is not sufficient. To be able to reduce the model it is also necessary that the inertial properties of both bodies, at $B_i$ and $B_j$, are expressed in orientations in which they are constant.

% -----------------------------------------------------------------------------------------------------------------
% -----------------------------------------------------------------------------------------------------------------
	\section{A low mobility mechanism example} \label{sec::the_system}
% -----------------------------------------------------------------------------------------------------------------
% -----------------------------------------------------------------------------------------------------------------

\begin{figure}[htb]	
	\begin{center}
		\includegraphics[width=0.45\textwidth]{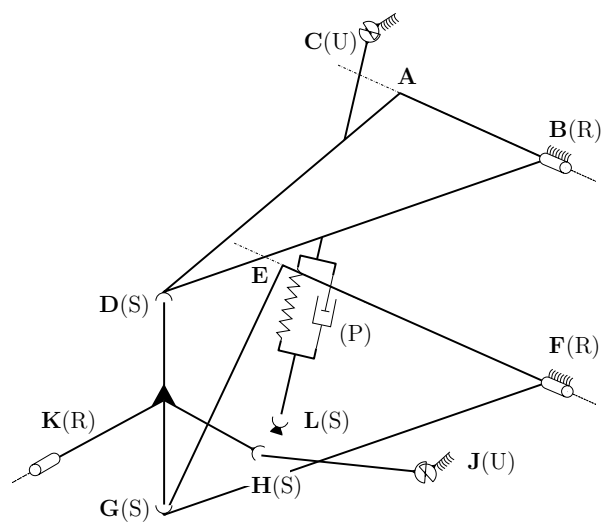}
		\caption{3D representation of the short-long arm (SLA) vehicle suspension. \textbf{B}(R), for instance, indicates that at point \textbf{B} a rotational (R) joint is present. Similarly, P, S and U stand for prismatic, spherical and universal, respectively.}
		\label{fig::suspension3d}
	\end{center}
\end{figure}

The previously described algorithms will be applied in this case to the SLA vehicle suspension shown in Fig.~\ref{fig::suspension3d}. The aim of this example is to show the drawbacks of using Gautiers algorithm with DP arithmetic and the advantages and shortcomings of the previously described alternatives based on the VP implementation and the symbolic method.

\begin{figure}[htb]	
	\begin{center}
		\includegraphics[width=0.48\textwidth]{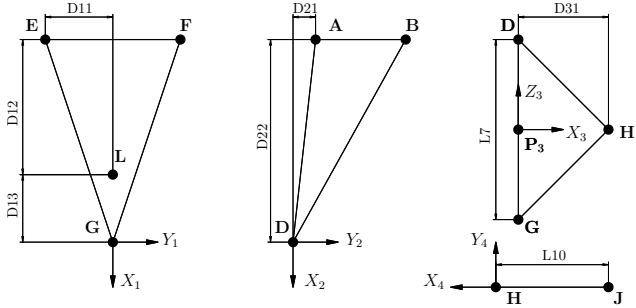}
		\caption{Projections of bodies 1, 2, 3 and 4 and definition of the geometric parameters}
		\label{fig::suspension2d}
	\end{center}
\end{figure}

The suspension is joined to the chassis (body 0) at points \textbf{B}, \textbf{F}, \textbf{C} and \textbf{J}, and segments $\overline{\textbf{AB}}$ and $\overline{\textbf{EF}}$ are parallel. The wheel (body 5, not represented in Fig.~\ref{fig::suspension3d}) is joined to the hub (body 3) at point \textbf{K}. The Y3 axis of the hub is coincident with the negative direction of the spinning axis of the wheel (X5). For the purpose of this paper, the chassis is assumed to be fixed to the ground. The system has 2 degrees of freedom and 3 closed loops. Geometric parameters $DKx$, $DKy$ and $DKz$ are the components of vector $\overline{DK}$ in the frame of reference of body 3. The rest of the geometric parameters that determine the location of the key points of the system are defined in Fig.~\ref{fig::suspension2d}. Bodies 6 and 7 are the 2 parts of the damper attached to solids 0 and 1, respectively.

This suspension can be considered to be a \emph{low mobility mechanism} since the amplitudes of the translations and rotations of the different bodies with respect to the chassis can be very small. For instance, the angular velocity of the hub can be any vector in a plane defined by vectors $\overline{\textbf{AB}}$ and $\overline{\textbf{DG}}$, but since the four bar sub-chain \textbf{BDGF} is almost a parallelogram, the movement of the hub will mainly be a translation without rotation. This low mobility implies that some dynamic parameters have very small (but not zero) influence on the kinetics of the mechanism.

% -----------------------------------------------------------------------------------------------------------------
% -----------------------------------------------------------------------------------------------------------------
	\section{Results}	 \label{sec::results}
% -----------------------------------------------------------------------------------------------------------------
% -----------------------------------------------------------------------------------------------------------------

The three methods to determine the base inertial parameters expressions described in Sections~\ref{sec::gautier} to~\ref{sec::symbolic_method} have been applied to the SLA suspension. 

% -----------------------------------------------------------------------------------------------------------------
\subsection{Numerical base inertial parameters of the SLA suspension}
\begin{figure}[htb]	
	\begin{center}
		\includegraphics[width=0.45\textwidth]{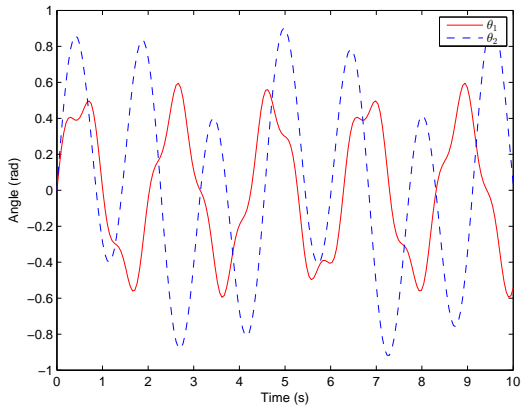}
		\caption{Excitation trajectories. $\theta_1$ and $\theta_2$ represent the rotation angle of body 1 with respect to the chassis, and the rotation angle of the wheel with respect to the hub}
		\label{fig::trajectories}
	\end{center}
\end{figure}

% -----------------------------------------------------------------------------------------------------------------
The base parameters expressions resulting from the application of Gautiers method to the SLA suspension, with DP and VP arithmetic, are respectively shown in Tables~\ref{tab::param_base_dp_mx5} and~\ref{tab::param_base_vpa}. 
The excitation trajectory used to evaluate matrix $W$ is shown in Fig.~\ref{fig::trajectories}. The trajectories have been designed as a finite Fourier Series of two harmonics (as in~\cite{Valero2013}) where the amplitudes have been selected to cover the full range of the motion of the suspension, leading to a better numerical conditioning of $W$.
The exciting trajectories for angles $\theta_1$ (rotation angle of body 1 with respect to the chassis) and $\theta_2$ (rotation angle of the wheel with respect to the hub) are:
\begin{subequations} \label{eq::trajectory}
  \begin{align}
  \theta_1&=0.5\sin(3.0\, t)\,\,\,\,\, + 0.1\sin(10.0\, t) \\
  \theta_2&=0.3\sin(\sqrt{2.0}\,t) +0.7\sin(\sqrt{17.0}\,t).
  \end{align}
\end{subequations}

\begin{center}
\begin{table*}[t]
\centering
\resizebox{0.8\textwidth}{!}{%
	\begin{tabular}{|l|l|l|}
	\hline
	\multicolumn{2}{|l|}{$\phi_{b,01}=mx_1 + 2.5\cdot Iyy_1 - 0.065\cdot m_7 + 4.6e10\cdot mx_5$}		\\
	\hline
	\multicolumn{2}{|l|}{$\phi_{b,02}=mz_1 - 4.3e4\cdot mx_5$}	\\
	\hline
	\multicolumn{2}{|l|}{$\phi_{b,03}=mx_2 + 4.0\cdot Iyy_2 - 1.6e9\cdot mx_5$}	\\
	\hline
	\multicolumn{2}{|l|}{$\phi_{b,04}=mz_2 + 3.1e4\cdot mx_5$}	\\
	\hline
	\multicolumn{2}{|l|}{$\phi_{b,05}=m_3 + m_1 - 16.0\cdot Iyy_2 - 9.8\cdot Iyy_4 - 6.2\cdot Iyy_1+ m_2 + m_4 + m_5 + 0.96\cdot m_7 - 9.6e10\cdot mx_5$}	\\
	\hline
	\multicolumn{2}{|l|}{$\phi_{b,06}=mx_3 + 0.13\cdot m_4 - 1.3\cdot Iyy_4 - 0.017\cdot m_5 + 3.3e9\cdot mx_5$}	\\
	\hline
	\multicolumn{2}{|l|}{$\phi_{b,07}=my_3 + 0.038\cdot Ixy_7 - 0.04\cdot Ixx_7 - 0.035\cdot Ixz_7 - 15.0\cdot Iyy_1 - 60.0\cdot Iyy_2 - 38.0\cdot Iyy_4 - 0.028\cdot Iyy_7$} \\
% 	\multicolumn{2}{|l|}{$\phantom{\phi_{b,07}=my_3}   $}\\
	\multicolumn{2}{|l|}{$\phantom{\phi_{b,07}=my_3} - 1.1\cdot Iyz_7 - 4.9\cdot Izz_5 + 1.1\cdot m_1+ 5.5\cdot m_2 + 4.0\cdot m_4 + 3.9\cdot m_5 + 2.3\cdot m_7 + 3.4e16\cdot mx_5$}	\\
	\hline
	\multicolumn{2}{|l|}{$\phi_{b,08}=mz_3 + 1.3\cdot Iyy_1 - 3.4\cdot Iyy_2 - 0.21\cdot m_1 + 0.21\cdot m_2 - 0.015\cdot m_5 - 0.2\cdot m_7 + 2.9e10\cdot mx_5$}	\\
% 	\multicolumn{2}{|l|}{$\phantom{\phi_{b,08}=mz_3} $}	\\
	\hline
	\multicolumn{2}{|l|}{$\phi_{b,09}=Ixx_3 + 0.24\cdot Iyy_1 + 1.4\cdot Iyy_2 + 1.4\cdot Iyy_4 + 0.038\cdot Iyz_7 + 1.2\cdot Izz_5 $}	\\
	\multicolumn{2}{|l|}{$\phantom{\phi_{b,09}=Ixx_3} - 0.15\cdot m_2 - 0.14\cdot m_4 - 0.14\cdot m_5 - 0.039\cdot m_7 - 1.2e15\cdot mx_5$}\\
% 	\multicolumn{2}{|l|}{$\phi_{b,09}=Ixx_3 + 0.24\cdot Iyy_1 + 1.4\cdot Iyy_2 + 1.4\cdot Iyy_4 + 0.038\cdot Iyz_7 + 1.2\cdot Izz_5 - 0.15\cdot m_2 - 0.14\cdot m_4 - 0.14\cdot m_5 - 0.039\cdot m_7 - 1.2e15\cdot mx_5$}	\\
	\hline
	\multicolumn{2}{|l|}{$\phi_{b,10}=Ixy_3 - 0.25\cdot Iyy_1 - 1.0\cdot Iyy_2 - 0.64\cdot Iyy_4 - 0.018\cdot Iyz_7 - 0.082\cdot Izz_5 + 0.019\cdot m_1$}	\\
% 	\multicolumn{2}{|l|}{$\phantom{\phi_{b,10}=Ixy_3}   $}\\
	\multicolumn{2}{|l|}{$\phantom{\phi_{b,10}=Ixy_3} + 0.092\cdot m_2 + 0.067\cdot m_4+ 0.066\cdot m_5 + 0.039\cdot m_7 + 5.7e14\cdot mx_5$}\\
	\hline
	\multicolumn{2}{|l|}{$\phi_{b,11}=Ixz_3 - 1.9e9\cdot mx_5$}	\\
	\hline
	\multicolumn{2}{|l|}{$\phi_{b,12}=Iyy_3 - 0.73\cdot Iyy_2 - 0.28\cdot Iyy_1 - 0.17\cdot Iyy_4 + 0.046\cdot m_1 + 0.046\cdot m_2 + 0.017\cdot m_4 + 0.044\cdot m_7 - 3.2e9\cdot mx_5$}	\\
% 	\multicolumn{2}{|l|}{$\phantom{\phi_{b,12}=Iyy_3}  $}\\
	\hline
	\multicolumn{2}{|l|}{$\phi_{b,13}=Iyz_3 + 0.059\cdot m_5 - 0.89\cdot Iyy_2 - 0.57\cdot Iyy_4 - 0.22\cdot Iyy_1 - 0.016\cdot Iyz_7 - 0.073\cdot Izz_5$}	\\
% 	\multicolumn{2}{|l|}{$\phantom{\phi_{b,12}=Iyy_3}   $}\\
	\multicolumn{2}{|l|}{$\phantom{\phi_{b,12}=Iyy_3} + 0.017\cdot m_1 + 0.082\cdot m_2+ 0.06\cdot m_4 + 0.035\cdot m_7 + 5.1e14\cdot mx_5$}\\
	\hline
	\multicolumn{2}{|l|}{$\phi_{b,14}=Izz_3 + 1.2\cdot Iyy_4 + 1.2\cdot Izz_5 - 0.13\cdot m_4 - 0.14\cdot m_5 + 0.53\cdot Iyy_1 + 2.1\cdot Iyy_2$}	\\
% 	\multicolumn{2}{|l|}{$\phantom{\phi_{b,14}=Izz_3}   $}\\
	\multicolumn{2}{|l|}{$\phantom{\phi_{b,14}=Izz_3} + 0.038\cdot Iyz_7 - 0.041\cdot m_1- 0.2\cdot m_2 - 0.083\cdot m_7 - 1.2e15\cdot mx_5$}\\
	\hline
	$\phi_{b,15}=mx_4 - 3.1\cdot Iyy_4 + 8.0e9\cdot mx_5$	& $\phi_{b,29}=Iyz_5 + 5.8\cdot mx_5$	\\
	\hline
	$\phi_{b,16}=my_4 - 6.0e4\cdot mx_5$	& $  \phi_{b,30}=mx_6 + 8.1e10\cdot mx_5$\\
	\hline
	$\phi_{b,17}=mz_4 + 6.8e3\cdot mx_5$	& $\phi_{b,31}=my_6 - 4.3e10\cdot mx_5$\\
	\hline
	$\phi_{b,18}=Ixx_4 - 1.2e11\cdot mx_5$	& $\phi_{b,32}=mz_6 - 8.4e11\cdot mx_5$\\
	\hline
	$\phi_{b,19}=Izz_4 - Iyy_4 - 3.1e10\cdot mx_5$	&$\phi_{b,33}=Ixx_6 + Ixx_7 + 4.1e12\cdot mx_5$\\
% 	\cline{1-1}
	\hline
	$\phi_{b,20}=Ixy_4 - 2.2e10\cdot mx_5$	& $\phi_{b,34}=Iyy_6 + Iyy_7 - 0.012\cdot Iyy_4 - 0.019\cdot Iyy_2 + 1.1e13\cdot mx_5$\\
	\hline
	$\phi_{b,21}=Ixz_4 + 1.5e10\cdot mx_5$	& $\phi_{b,35}=Izz_6 + Izz_7 - 5.7e9\cdot mx_5$\\
	\hline
	$\phi_{b,22}=Iyz_4 - 7.3e8\cdot mx_5$	& $\phi_{b,36}=Ixy_6 + Ixy_7 - 0.014\cdot Iyy_2 + 8.0e12\cdot mx_5$\\
	\hline
	\multicolumn{2}{|l|}{$\phi_{b,23}=my_5 + 0.04\cdot Ixx_7 - 0.038\cdot Ixy_7 + 0.035\cdot Ixz_7 + 15.0\cdot Iyy_1 + 60.0\cdot Iyy_2 + 38.0\cdot Iyy_4 + 0.028\cdot Iyy_7 $}	\\
% 	\multicolumn{2}{|l|}{$\phantom{\phi_{b,23}=my_5}  $}	\\
	\multicolumn{2}{|l|}{$\phantom{\phi_{b,23}=my_5} + 1.1\cdot Iyz_7 + 4.9\cdot Izz_5 - 1.1\cdot m_1- 5.5\cdot m_2 - 4.0\cdot m_4 - 3.9\cdot m_5 - 2.3\cdot m_7 - 3.4e16\cdot mx_5$}\\
	\hline
	$\phi_{b,24}=mz_5 - 3.6\cdot mx_5$	& $\phi_{b,37}=Ixz_6 + Ixz_7 + 5.6e10\cdot mx_5$\\
	\hline
	$\phi_{b,25}=Ixx_5 - Izz_5 - 46.0\cdot mx_5$ &	$\phi_{b,38}=Iyz_6 + Iyz_7 - 3.0e11\cdot mx_5$\\
	\hline
	$\phi_{b,26}=Iyy_5 + 1.3\cdot mx_5$	& $\phi_{b,39}=mx_7 - 8.1e10\cdot mx_5$\\
	\hline
	$\phi_{b,27}=Ixy_5 + 3.3\cdot mx_5$	& $\phi_{b,40}=my_7 + 4.3e10\cdot mx_5 $\\
	\hline
	$\phi_{b,28}=Ixz_5 - 8.2\cdot mx_5$	& $\phi_{b,41}=mz_7 + 8.4e11\cdot mx_5$\\
	\hline
	\end{tabular}
	}
	\caption{Numerical DP Base Inertial Parameters of the SLA suspension (D22 = 0.2501, D12= 0.3196, D13 = 0.082, D31 = 0.1301, L7 = 0.427, L10 = 0.319). The elements of $\beta<10^{-2}$ have been neglected.}
	\label{tab::param_base_dp_mx5}
\end{table*}
\end{center}
\begin{center}
\begin{table*}[t]
\centering
\resizebox{0.8\textwidth}{!}{%
	\begin{tabular}{|l|l|l|}
	\hline
	\multicolumn{3}{|l|}{$\phi_{b,01}=mx_1+2.4900398\cdot Iyy_1-0.065256972\cdot m_7$}	\\
	\hline
	\multicolumn{3}{|l|}{$\phi_{b,02}=mz_1$}	\\
	\hline
	\multicolumn{3}{|l|}{$\phi_{b,03}=mx_2+3.9984006\cdot Iyy_2$}	\\
	\hline
	\multicolumn{3}{|l|}{$\phi_{b,04}=mz_2$}	\\
	\hline
	\multicolumn{3}{|l|}{$\phi_{b,05}=m_3+m_1-15.987208\cdot Iyy_2-9.8269475\cdot Iyy_4-6.2002984\cdot Iyy_1+m_2+m_4+m_5+0.95830919\cdot m_7$}	\\
	\hline
	\multicolumn{3}{|l|}{$\phi_{b,06}=mx_3+0.1301\cdot m_4-1.2784859\cdot Iyy_4-0.0168\cdot m_5$}	\\
	\hline
	\multicolumn{3}{|l|}{$\phi_{b,07}=my_3-0.018\cdot m_5-mx_5$}	\\
	\hline
	\multicolumn{3}{|l|}{$\phi_{b,08}=mz_3+1.3237637\cdot Iyy_1-3.4132688\cdot Iyy_2-0.2135\cdot m_1+0.2135\cdot m_2-0.015\cdot m_5-0.20459901\cdot m_7$}	\\
	\hline
	\multicolumn{3}{|l|}{$\phi_{b,09}=Ixx_3+0.04558225\cdot m_2-0.7287329\cdot Iyy_2+0.04558225\cdot m_1+0.043681889\cdot m_7$}	\\
	\multicolumn{3}{|l|}{$\phantom{\phi_{b,09}=Ixx_3}-0.28262355\cdot Iyy_1+Izz_5+0.000549\cdot m_5+0.036\cdot mx_5$}	\\
	\hline
	\multicolumn{3}{|l|}{$\phi_{b,10}=Ixy_3-0.0003024\cdot m_5-0.0168\cdot mx_5$}	\\
	\hline
	\multicolumn{3}{|l|}{$\phi_{b,11}=Ixz_3-0.000252\cdot m_5$}	\\
	\hline
	\multicolumn{3}{|l|}{$\phi_{b,12}=Iyy_3+0.04558225\cdot m_2-0.7287329\cdot Iyy_2+0.04558225\cdot m_1+0.043681889\cdot m_7$}	\\
	\multicolumn{3}{|l|}{$\phantom{\phi_{b,12}=Iyy_3}-0.28262355\cdot Iyy_1+0.01692601\cdot m_4-0.16633101\cdot Iyy_4+0.00050724\cdot m_5$}\\
	\hline
	\multicolumn{3}{|l|}{$\phi_{b,13}=Iyz_3-0.00027\cdot m_5-0.015\cdot mx_5$}	\\
	\hline
	\multicolumn{3}{|l|}{$\phi_{b,14}=Izz_3-0.16633101\cdot Iyy_4+Izz_5+0.01692601\cdot m_4+0.00060624\cdot m_5+0.036\cdot mx_5$}	\\
	\hline
	$\phi_{b,15}=mx_4-3.1347962\cdot Iyy_4$	\phantom{0000} 		&	$\phi_{b,24}=mz_5$	&	$\phi_{b,33}=Ixx_6+Ixx_7$	\\
	\hline
	$\phi_{b,16}=my_4$						&	$\phi_{b,25}=Ixx_5-Izz_5$\phantom{00000000}&	$\phi_{b,34}=Iyy_6+Iyy_7$	\\
	\hline
	$\phi_{b,17}=mz_4$						&	$\phi_{b,26}=Iyy_5$			&	$\phi_{b,35}=Izz_6+Izz_7$	\\
	\hline
	$\phi_{b,18}=Ixx_4$						&	$\phi_{b,27}=Ixy_5$			&	$\phi_{b,36}=Ixy_6+Ixy_7$	\\
	\hline
	$\phi_{b,19}=Ixy_4$						&	$\phi_{b,28}=Ixz_5$			&	$\phi_{b,37}=Ixz_6+Ixz_7$	\\
	\hline
	$\phi_{b,20}=Ixz_4$						&	$\phi_{b,29}=Iyz_5$			&	$\phi_{b,38}=Iyz_6+Iyz_7$	\\
	\hline
	$\phi_{b,21}=Iyz_4$						&	$\phi_{b,30}=mx_6$			&	$\phi_{b,39}=mx_7$			\\
	\hline
	$\phi_{b,22}=Izz_4-Iyy_4$					&	$\phi_{b,31}=my_6$			&	$\phi_{b,40}=my_7$			\\
	\hline
	$\phi_{b,23}=my_5$						&	$\phi_{b,32}=mz_6$			&	$\phi_{b,41}=mz_7$			\\
	\hline
	\end{tabular}
	}
	\caption{Numerical VP Base Inertial Parameters of the SLA suspension (D22 = 0.2501, D12= 0.3196, D13 = 0.082, D31 = 0.1301, L7 = 0.427, L10 = 0.319).}
	\label{tab::param_base_vpa}
\end{table*}
\end{center}	

Almost all the elements of $\beta$ of the DP calculations are actually different from zero. To make Tab.~\ref{tab::param_base_dp_mx5} readable, the $\beta$-s whose absolute value was smaller than $10^{-2}$ have been removed.

% %%%%%%%%%%%%%%%% begin figure %%%%%%%%%%%%%%%%%%%
% %%% the maximum width in double column is 6.85in
% \begin{figure*} 
% \centerline{\psfig{figure=figure/FMANU_MD_04_1274_13.ps,width=6.85in}}
% \caption{A figure expanded to double column width the text from Figure~\ref{fig_example3.ps}}
% \label{fig_example4.ps}
% \end{figure*}
% %%%%%%%%%%%%%%%% end figure %%%%%%%%%%%%%%%%%%%

% -----------------------------------------------------------------------------------------------------------------
\subsection{Symbolic base inertial parameters of the SLA suspension}
% -----------------------------------------------------------------------------------------------------------------

In this Section the base inertial parameters expressions of the suspension are determined using the method of Ros \etal~\cite{Ros2012} previously described in Section~\ref{sec::symbolic_method}.

\subsubsection{Inertial parameters with no effect on the kinetics}
% -----------------------------------------------------------------------------------------------------------------
As described in Section~\ref{sec:non_affecting_inertial_properties}, the movement limitations of a certain body with respect to the ground, can make some of its parameters disappear from the equations of motion of the mechanism. Applying the 6 conditions of Section~\ref{sec:non_affecting_inertial_properties} to the 7 bodies of the mechanism, it can be concluded that the 13 parameters given in Tab.~\ref{tab::eliminated_parameters} will not appear in the equations of motion.

As an illustration, for the first row of Tab.~\ref{tab::eliminated_parameters}, since the direction of $\vOmega_1$ will always be parallel to the Y1 axis, the dipole of body 1 in that direction ($my_1$) will not appear in the equations of motion.

\subsubsection{Inertial parameters eliminated in the transfers}
% -----------------------------------------------------------------------------------------------------------------

As explained in Section~\ref{sec:inertia_transfers}, the relative movement limitations between a pair of bodies makes it possible to transfer a multipole between them without changing the Lagrangian of the mechanism. In practice, for almost any kinematic joint, a multipole transfer can be performed between the joined bodies.  Applying condition 6 of Section~\ref{sec:inertia_transfers} to spherical or universal joints, the monopole (the mass) can be transferred from one body to the other. Likewise, if two bodies are joined by a revolute joint, a monopole, a dipole and a quadrupole can be transferred.

For each transfers, the value of the transferred multipole is chosen so that the value of a parameter of the bodies involved is zero. In this way, the body (and the whole mechanism) will depend on one parameter less. Choosing proper values for the transferred multipoles, the base parameters expressions are obtained.

Tab.~\ref{tab::transferred_parameters} shows the inertial parameters that will be eliminated for each joint in the system, in total 16. Therefore, together with Tab.~\ref{tab::eliminated_parameters}, even without determining the symbolic base parameters expressions, it is straightforward to see that the number of base parameters for this system will be: 70 - 13 - 16 = 41.

After all the multipole transfers have been made, the resulting base parameters expressions are listed in Tab.~\ref{tab::tabla_parametros_base}. The thorough determination of the symbolic expressions has been deferred to the Appendix.

\renewcommand{\arraystretch}{1.2}

\begin{center}

\begin{table}[!htbp]
	\begin{center}
	\begin{tabular}{|c|c|l|}
	\hline
	\textbf{Body}		&	\textbf{Condition} 	&	\textbf{Inertial Parameters}							\\
	\hline
	\hline
	1			&		2		&	$my_1$										\\
	\hline
	1			&		5		&	$Ixx_1$, $Ixy_1$, $Ixz_1$, $Iyz_1$, $Izz_1$ 	\\
	\hline
	2			&		2		&	$my_2$										\\
	\hline
	2			&		5		&	$Ixx_2$, $Ixy_2$, $Ixz_2$, $Iyz_2$, $Izz_2$ 	\\
	\hline
	3,4,5,7			&		--		&	--											\\
	\hline
	6			&		1		& 	$m_6$										\\
	\hline
	\end{tabular}
\end{center}	\caption{Solid number, condition satisfied, and inertial parameters that do not appear in the equations of motion}	\label{tab::eliminated_parameters}
\end{table}
\end{center}	

\renewcommand{\arraystretch}{1}

\renewcommand{\arraystretch}{1.2}

\begin{center}

\begin{table}[!htbp]
% 	\begin{center}
\resizebox{0.49\textwidth}{!}{%

	\begin{tabular}{|c|c|c|c|c|}
	\hline
% 	\textbf{Bodies} & \textbf{Joint Type} & \textbf{Condition} & \textbf{Transferred Multipole} & \textbf{Eliminated Parameter} \\
	\textbf{Bodies} & \textbf{Joint} & \textbf{Condition} & \textbf{Transferred} & \textbf{Eliminated} \\
	\phantom{Bodies} & \textbf{Type} & \phantom{Condition} & \textbf{Multipole} & \textbf{Parameter} \\
	\hline	
	\hline
	1-7 	&	S	&	1	&	$m$						&	$m_7$	\\
	\hline
	0-1	&	R	&	1	&	$m$						&	$Iyy_1$	\\
	\hline
	0-2	&	R	&	1	&	$m$						&	$Iyy_2$	\\
	\hline
	0-4	&	U	&	1	&	$m$						&	$Iyy_4$	\\
	\hline
	1-3 	&	S	&	1	&	$m$						&	$m_1$	\\
	\hline
	2-3 	&	S	&	1	&	$m$						&	$m_2$	\\
	\hline
	5-3 	&	R	& 1,2,5	&	$m$, $\mathbf{d}$ and $\overline{\overline{\mathbf{q}}}$ 	&	$m_5, mx_5, Izz_5$	 \\
	\hline
	4-3 	&	S	&	1	&	$m$						&	$m_4$	\\
	\hline
	6-7 	&	P	&	6	&	all $\overline{\overline{\mathbf{q}}}_i$		&	$\boldsymbol{I}_7$	\\
	\hline
	\end{tabular}
% \end{center}	
}
\caption{Eliminated Parameters and Transferred Multipoles for each joint of the suspension}		\label{tab::transferred_parameters}
\end{table}
\end{center}	

\renewcommand{\arraystretch}{1}

\renewcommand{\arraystretch}{1.2}

\begin{center}
\begin{table*}[t]
\centering
\resizebox{0.8\textwidth}{!}{%
	\begin{tabular}{|l|l|l|}
	\hline
	\multicolumn{3}{|l|}{$\phi_{b,01}= mx_1-m_7\cdot D13 +\frac{Iyy_1+m_7\cdot D13^2}{D12+D13}$}	\\
	\hline
	\multicolumn{3}{|l|}{$\phi_{b,02}= mz_1 $}	\\
	\hline
	\multicolumn{3}{|l|}{$\phi_{b,03}= mx_2 +\frac{Iyy_2}{D22}$}	\\
	\hline
	\multicolumn{3}{|l|}{$\phi_{b,04}= mz_2 $}	\\
	\hline
	\multicolumn{3}{|l|}{$\phi_{b,05}= m_3+m_1+m_7\cdot (1-\frac{D13^2}{(D12+D13)^2})-\frac{Iyy_1}{(D12+D13)^2}+m_2-\frac{Iyy_2}{D22^2}+m_4 -\frac{Iyy_4}{L10^2}+m_5$}	\\
	\hline
	\multicolumn{3}{|l|}{$\phi_{b,06}= mx_3 +m_5 \cdot DKx +(m_4-\frac{Iyy_4}{L10^2})D31$}	\\
	\hline
	\multicolumn{3}{|l|}{$\phi_{b,07}= my_3 +m_5 \cdot DKy - mx_5 $}	\\
	\hline
	\multicolumn{3}{|l|}{$\phi_{b,08}= mz_3 +m_5 \cdot (DKz+\frac{L7}{2})+(m_2-\frac{Iyy_2}{D22^2}-m_1)\cdot \frac{L7}{2}- m_7\cdot (1-\frac{D13^2}{(D12+D13)^2})\cdot \frac{L7}{2}+\frac{Iyy_1}{(D12+D13)^2} \cdot \frac{L7}{2}$}	\\
	\hline
% 	\multicolumn{3}{|l|}{$\phi_{b,09}= Ixx_3 + (m_2-\frac{Iyy_2}{D22^2}+m_1+m_7-\frac{Iyy_1+m_7\cdot D13^2}{(D12+D13)^2} )(\frac{L7}{2})^2 + Izz_5 $} \\
% 	\multicolumn{3}{|l|}{$\phantom{\phi_{b,09}= Ixx_3}+m_5\cdot (DKy^2+(DKz+\frac{L7}{2})^2)-2\cdot DKy\cdot mx_5$}	\\
% 	\hline
	\multicolumn{3}{|l|}{$\phi_{b,09}= Ixx_3 + (m_2-\frac{Iyy_2}{D22^2}+m_1+m_7\cdot (1-\frac{D13^2}{(D12+D13)^2})-\frac{Iyy_1}{(D12+D13)^2} )(\frac{L7}{2})^2 $} \\
	\multicolumn{3}{|l|}{$\phantom{\phi_{b,09}= Ixx_3}+ Izz_5 +m_5\cdot (DKy^2+(DKz+\frac{L7}{2})^2)-2\cdot DKy\cdot mx_5$}	\\
	\hline
	\multicolumn{3}{|l|}{$\phi_{b,10}= Ixy_3 -DKx\cdot DKy\cdot m_5+DKx\cdot mx_5$}	\\
	\hline
	\multicolumn{3}{|l|}{$\phi_{b,11}= Ixz_3 -DKx\cdot(DKz+\frac{L7}{2})\cdot m_5$}	\\
	\hline
% 	\multicolumn{3}{|l|}{$\phi_{b,12}= Iyy_3 +(m_2-\frac{Iyy_2}{D22^2}+m_1+m_7-\frac{Iyy_1+m_7\cdot D13^2}{(D12+D13)^2})(\frac{L7}{2})^2$} \\
% 	\multicolumn{3}{|l|}{$\phantom{\phi_{b,12}= Iyy_3}+ (m_4- \frac{Iyy_4}{L10^2}) D31^2+ (DKx^2+(DKz+\frac{L7}{2})^2)\cdot m_5$}	\\
% 	\hline
	\multicolumn{3}{|l|}{$\phi_{b,12}= Iyy_3 +(m_2-\frac{Iyy_2}{D22^2}+m_1+m_7\cdot (1-\frac{D13^2}{(D12+D13)^2})-\frac{Iyy_1}{(D12+D13)^2})(\frac{L7}{2})^2$} \\
	\multicolumn{3}{|l|}{$\phantom{\phi_{b,12}= Iyy_3}+ (m_4- \frac{Iyy_4}{L10^2}) D31^2+ (DKx^2+(DKz+\frac{L7}{2})^2)\cdot m_5$}	\\
	\hline
	\multicolumn{3}{|l|}{$\phi_{b,13}= Iyz_3 -m_5\cdot DKy\cdot(DKz+\frac{L7}{2})+(DKz+\frac{L7}{2})\cdot mx_5$}	\\
	\hline
	\multicolumn{3}{|l|}{$\phi_{b,14}= Izz_3 +Izz_5 + m_5\cdot(DKx^2+DKy^2)+(m_4-\frac{Iyy_4}{L10^2}) D31^2 - 2\cdot DKy\cdot mx_5$}	\\
% 	\hline
	\hline
	$\phi_{b,15}= mx_4 - \frac{Iyy_4}{L10}$	\phantom{0000000000}	&	$\phi_{b,24}= mz_5$\phantom{00000000000000}	&	$\phi_{b,33}= Ixx_6+Ixx_7$	\\
	\hline
	$\phi_{b,16}= my_4 $						&	$\phi_{b,25}= Ixx_5$				&	$\phi_{b,34}= Ixy_6+Ixy_7$	\\
	\hline
	$\phi_{b,17}= mz_4$						&	$\phi_{b,26}= Ixy_5$				&	$\phi_{b,35}= Ixz_6+Ixz_7$	\\
	\hline
	$\phi_{b,18}= Ixx_4$						&	$\phi_{b,27}= Ixz_5$				&	$\phi_{b,36}= Iyy_6+Iyy_7$	\\
	\hline
	$\phi_{b,19}= Ixy_4$						&	$\phi_{b,28}= Iyy_5-Izz_5$			&	$\phi_{b,37}= Iyz_6+Iyz_7$	\\
	\hline
	$\phi_{b,20}= Ixz_4$						&	$\phi_{b,29}= Iyz_5$				&	$\phi_{b,38}= Izz_6+Izz_7$	\\
	\hline
	$\phi_{b,21}= Iyz_4$						&	$\phi_{b,30}= mx_6$				&	$\phi_{b,39}= mx_7$		\\
	\hline
	$\phi_{b,22}= Izz_4-Iyy_4$					&	$\phi_{b,31}= my_6$				&	$\phi_{b,40}= my_7$		\\
	\hline
	$\phi_{b,23}= my_5$						&	$\phi_{b,32}= mz_6$				&	$\phi_{b,41}= mz_7$		\\
	\hline
	\end{tabular}
	}
	\caption{Symbolic Base Inertial Parameters of the SLA suspension}	\label{tab::tabla_parametros_base}
\end{table*}
\end{center}

\renewcommand{\arraystretch}{1}

% -----------------------------------------------------------------------------------------------------------------
% -----------------------------------------------------------------------------------------------------------------
	\section{Discussion}	\label{sec::discussion}
% -----------------------------------------------------------------------------------------------------------------
% -----------------------------------------------------------------------------------------------------------------

In the algorithm of Gautier~\cite{Gautier1991}, the rank of the observation matrix is calculated numerically as the number of singular values that are larger than a certain machine dependent tolerance. It is common to optimize the set of points used to assemble the observation matrix so that its numerical conditioning is improved making it possible to determine the rank, and therefore the number of base parameters. For mechanisms with low mobility, using DP arithmetic there is often no natural cut-off value that clearly separates the singular values that are larger or smaller than the given tolerance.
In this situation, the rank obtained depends on truncation errors and therefore will be calculated incorrectly, making Gautier's procedure unsuitable. Moreover, without the rank information it will not be possible to optimize a trajectory to properly excite the mechanism since all common optimization criteria depend on the rank of $W$~\cite{Sun2008}.

\begin{figure}[htb]	
	\begin{center}
		\includegraphics[width=0.45\textwidth]{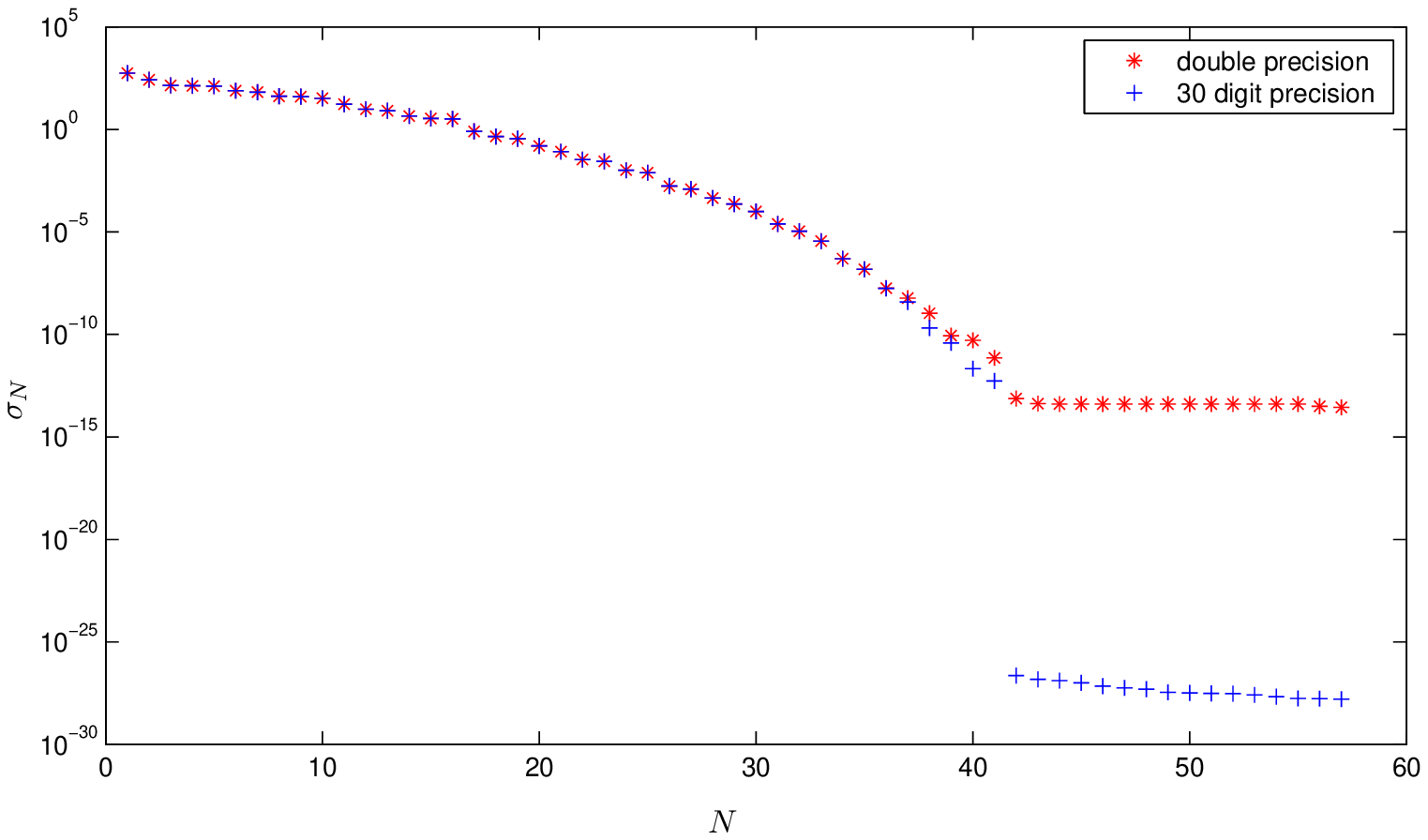}
		\caption{Singular values for $W$: VP with 30 digits ($+$) and DP ($*$)}
		\label{fig::sigma_VPA_double}
	\end{center}
\end{figure}

To clearly illustrate these problems, Fig.~\ref{fig::sigma_VPA_double} represents the singular values of $W$ using DP and VP arithmetic (using 30 digit precision\footnotemark). 
The expected ridge after the $41^{st}$ singular value is clearly observable in the case of VP arithmetic, but less so in the case of using DP. Moreover, the differences between calculations of the singular values from the $36^{th}$ to the $41^{st}$ with DP and VP arithmetic show that those using DP are not accurate at all.

\footnotetext{VP calculations with more digits further reduce the singular values after the $41^{st}$.}

If the rank of matrix $W$ is not known the base inertial parameters expressions will be calculated erroneously. Moreover, even if a correct rank is chosen to calculate the base parameters expressions using DP, the numerical values obtained for $\beta$ are markedly different from the correct ones.

These expressions can be compared to the corresponding parameters in Tab.~\ref{tab::param_base_vpa}, where the base parameters expressions of the SLA suspension have been calculated using the VP-based Gautier's algorithm.
As an illustration, the first base parameter ($\phi_{b1}$) can be compared. The obtained DP expression for this parameter is:

\begin{equation}
	\begin{split}
		\phi_{b,01}=mx_1&+ 2.5\cdot Iyy_1         \phantom{!}\quad -0.065 \cdot m_7 \quad\phantom{!} - \mathbf{4.6e^{11}\cdot mx_5}\\
				& + 2.0 e^{-6}\cdot Ixy_7 \, - 1.2e^{-7}\cdot Ixx_7 \, - 2.7e^{-6}\cdot Ixz_7 \\
				& - 6.4e^{-4}\cdot Iyy_2  \, - 6.9e^{-4}\cdot Iyy_4 \, + 1.1e^{-7}\cdot Iyy_7 \\
				& - 1.8e^{-6}\cdot Iyz_7  \, - 2.2e^{-5}\cdot Izz_5 \, - 6.8e^{-6}\cdot Izz_7 \\
				& + 6.4e^{-5}\cdot m_1 \phantom{a} \, + 5.4e^{-5}\cdot m_2 \phantom{a!} \, + 4.4e^{-5}\cdot m_4 \phantom{!!} \\
				& + 7.8e^{-5}\cdot m_5.
	\end{split}	
\end{equation}

While the values multiplying $Iyy_1$ and $m_7$ are close to those obtained with VP arithmetic, the term multiplying $mx_5$ is completely out of range. Moreover, many other terms appear that should not be present as the symbolic or VP calculations demonstrate. This clearly shows that applying Gautier's algorithm using DP can give incorrect results in the case of low mobility mechanisms.

The symbolic base parameters expressions shown in Tab.~\ref{tab::tabla_parametros_base} have been numerically evaluated, and they have been found to be identical to those in Tab.~\ref{tab::param_base_vpa}, up to the employed precision, proving the correctness and suitability of the symbolic and VP arithmetic based numerical algorithms.

As has been shown, the implementation of Gautier's algorithm using VP with a large enough number of digits can be used to correctly determine the base parameters expressions of low mobility mechanisms.
Two VP implementations of the algorithm have been employed: the first one using  MATLAB$^{TM}$ Symbolic Toolbox, and the second one Maple$^{TM}$. For the SLA suspension analyzed, the MATLAB based implementation can take more than one hour for a single run of the VP based Gautier's algorithm using a moderate number of digits, while Maple takes 5 minutes even using 500 digits. These computational requirements render these algorithms unsuitable for many problems of interest (model reduction and mechanism optimization, among others). For these problems, the availability of symbolic expressions of the base parameters will greatly reduce the required computational time.

% -----------------------------------------------------------------------------------------------------------------
% -----------------------------------------------------------------------------------------------------------------
	\section{Conclusions}
% -----------------------------------------------------------------------------------------------------------------
% -----------------------------------------------------------------------------------------------------------------

In this paper three methods for the calculation of the base inertial parameters expressions have been reviewed. Additionally, they have been applied in detail to a low mobility mechanism and the results have been compared.

It has been shown that Gautiers standard numerical algorithm can fail to calculate correct base parameters expressions when the mechanism at hand has low mobility. 
This is due to the lack of a ridge in the singular values of the observation matrix which can lead to an erroneous rank calculation.
In this situation, continuing with Gautiers algorithm is impossible, and even selecting the correct rank of the observation matrix the obtained results can be erroneous.
Moreover, trajectory optimization criteria are not well defined when the observation matrix is close to be rank deficient.

A VP implementation of Gautiers algorithm has been proposed for the first time, and it has revealed to be a suitable method to calculate the base inertial parameters expressions.
Some important details of a successful VP implementation have been described.
The computational time of the VP algorithm together with the multiple executions needed to prove convergence make such approach prohibitively expensive for some applications.

The application of the reviewed symbolic method to the low mobility example has been presented in detail in the appendix.
The symbolic method has shown to fully agree with the VP implementation, but with the benefit of not requiring an exciting trajectory, and providing symbolic expressions of the base inertial parameters This makes them more suitable for a wide range of applications, such as model reduction and mechanism optimization.
Additionally, symbolic expressions also provide a deeper insight into the kinetics of the mechanism.

\begin{acknowledgment}
This paper has been possible owing to the funding from the Science and Innovation Ministry of the Spanish Government by means of the Research and Technological Development Project DPI2010-20814-C02-01 (IDEMOV).
\end{acknowledgment}

% Here's where you specify the bibliography style file.
% The full file name for the bibliography style file 
% used for an ASME paper is asmems4.bst.
\bibliographystyle{asmems4}

% Here's where you specify the bibliography database file.
% The full file name of the bibliography database for this
% article is asme2e.bib. The name for your database is up
% to you.
\bibliography{bibliografia}

% -----------------------------------------------------------------------------------------------------------------
% -----------------------------------------------------------------------------------------------------------------
	\appendix
	\section*{Appendix A: Determination of expressions for the base inertial parameters}
% -----------------------------------------------------------------------------------------------------------------
% -----------------------------------------------------------------------------------------------------------------

In this Appendix the symbolic expressions of the base inertial parameters of the SLA suspension will be determined using the algorithm of Ros \etal~\cite{Ros2012} reviewed in Section~\ref{sec::symbolic_method}.
Since the parameters that will not appear in the equations of motion have already been determined in Section~\ref{sec:non_affecting_inertial_properties}, let us determine the symbolic base parameters expressions of the suspension making use of the 6 multipole-transfer conditions set in Section \ref{sec:inertia_transfers}.

\begin{itemize}
	\item \textbf{Joint 1-7}: Condition 1 is satisfied and $m_7$ is transferred from body 7 to body 1.
	\begin{itemize}
		\item $m_7'=0$
		\item $m_1'=m_1+m_7$
		\item $mx_1'=mx_1-m_7\cdot D13$ \\		
		\item $my_1'=my_1$
		\item $mz_1'=mz_1$
		\item $I_1'=I_1+m_7\cdot\begin{bmatrix}
		                   	0	&	0	&	0	\\
					0	&	D13^2	&	0	\\
					0	&	0	&	D13^2
		                   \end{bmatrix}$\\
		\phantom{-}\\
		Parameter $my_1'$ will not appear in the equations of motion, as shown in Tab.~\ref{tab::eliminated_parameters}.
		In particular, the only second moment of inertia that will not be eliminated in body 1 will be $Iyy_1'=Iyy_1+m_7\cdot D13^2$.
	\end{itemize}

	\item \textbf{Joint 0-1}: Condition 1 is satisfied and a mass ($m_{01}$) is transferred from body 0 to 1 at the intersecting point between segment $EF$ and axis X1. Since in a further step the mass of body 1 will be transferred to body 3 through the relative spherical joint, instead of eliminating $m_1$, the second moment of inertia $Iyy_1'$ will be eliminated.
    \begin{itemize}
% 		\item $Iyy_1''=Iyy_1'+m_{01}\cdot(D12+D13)^2, \quad (Iyy_1''=0)\Rightarrow \quad m_{01}=-\frac{Iyy_1'}{(D12+D13)^2}=-\frac{Iyy_1+m_7\cdot D13^2}{(D12+D13)^2}$
		\item $Iyy_1''=Iyy_1'+m_{01}\cdot(D12+D13)^2,$
		\item $(Iyy_1''=0)\Rightarrow m_{01}=-\frac{Iyy_1'}{(D12+D13)^2}=-\frac{Iyy_1+m_7\cdot D13^2}{(D12+D13)^2}$
		\item $m_1''=m_1'+m_{01}=m_1+m_7-\frac{Iyy_1+m_7\cdot D13^2}{(D12+D13)^2}$
% 		\item $mx_1''=mx_1'-m_{01}\cdot (D12+D13)=mx_1-m_7\cdot D13+\frac{Iyy_1+m_7\cdot D13^2}{(D12+D13)}$
		\item $mx_1''=mx_1'-m_{01}\cdot (D12+D13)$
		\item $\phantom{mx_1''}=mx_1-m_7\cdot D13+\frac{Iyy_1+m_7\cdot D13^2}{(D12+D13)}$
		\item $mz_1''=mz_1$
    \end{itemize}

	\item \textbf{Joint 0-2}: Analogous to Joint 0-1.
    \begin{itemize}
% 		\item $Iyy_2'=Iyy_2+m_{02}\cdot D22^2, \quad (Iyy_2'=0)\Rightarrow \quad m_{02}=-\frac{Iyy_2}{D22^2}$
		\item $Iyy_2'=Iyy_2+m_{02}\cdot D22^2,$
		\item $(Iyy_2'=0)\Rightarrow m_{02}=-\frac{Iyy_2}{D22^2}$
		\item $m_2'=m_2+m_{02}=m_2-\frac{Iyy_2}{D22^2}$
		\item $mx_2'=mx_2-m_{02}\cdot D22=mx_2+\frac{Iyy_2}{D22}$
		\item $mz_2'=mz_2$
    \end{itemize}

	\item \textbf{Joint 0-4}: Condition 1 is satisfied, and a mass is transferred from body 0 to point J of body 4, $m_{04}$. With this mass, the second moment of inertia $Iyy_4$ is eliminated.
    \begin{itemize}
% 		\item $Iyy_4'=Iyy_4+m_{04}\cdot L10^2, \quad (Iyy_4'=0)\Rightarrow \quad m_{04}=-\frac{Iyy_4}{L10^2}$
		\item $Iyy_4'=Iyy_4+m_{04}\cdot L10^2,$
		\item $(Iyy_4'=0)\Rightarrow m_{04}=-\frac{Iyy_4}{L10^2}$
		\item $Izz_4'=Izz_4+m_{04}\cdot L10^2=Izz_4- Iyy_4$
		\item $m_4'=m_4+m_{04}=m_4-\frac{Iyy_4}{L10^2}$
		\item $mx_4'=mx_4+m_{04}\cdot L10=mx_4-\frac{Iyy_4}{L10}$
		\item $my_4'=my_4$
		\item $mz_4'=mz_4$
    \end{itemize}

	\item \textbf{Joint 1-3}: Condition 1 is satisfied and the mass of body 1 is transferred to body 3.
	\begin{itemize}
		\item $m_1'''=0$ \\
		\item $m_3'=m_3+m_1''=m_3+m_1+m_7-\frac{Iyy_1+m_7\cdot D13^2}{(D12+D13)^2}$ \\
		\item $mz_3'=mz_3-m_1''\cdot\frac{L7}{2}$
		\item $\phantom{mz_3'}=mz_3-(m_1+m_7-\frac{Iyy_1+m_7\cdot D13^2}{(D12+D13)^2})\cdot \frac{L7}{2}$\\
		\item $I_3'=I_3+m_1''\cdot\begin{bmatrix}
                   	(\frac{L7}{2})^2	&	0			&	0	\\
			0			&	(\frac{L7}{2})^2	&	0	\\
			0			&	0			&	0
                   \end{bmatrix}$	\\
		\phantom{-}\\
		In particular, the second moments of inertia of body 3 will be:
		\item $Ixx_3'=Ixx_3+(m_1+m_7-\frac{Iyy_1+m_7\cdot D13^2}{(D12+D13)^2})\cdot (\frac{L7}{2})^2$
		\item $Iyy_3'=Iyy_3+(m_1+m_7-\frac{Iyy_1+m_7\cdot D13^2}{(D12+D13)^2})\cdot (\frac{L7}{2})^2$
		\item $Izz_3'=Izz_3$
	\end{itemize}

	\item \textbf{Joint 2-3}: Condition 1 is satisfied and the complete mass of body 2 is transferred to body 3.
	\begin{itemize}
		\item $m_2''=0$ \\
		\item $m_3''=m_3'+m_2'$
		\item $\phantom{m_3''}=m_3+m_1+m_7-\frac{Iyy_1+m_7\cdot D13^2}{(D12+D13)^2}+m_2-\frac{Iyy_2}{D22^2}$ \\
		\item $mz_3''=mz_3'+m_2'\cdot\frac{L7}{2}$
		\item $\phantom{}=mz_3+(m_2-\frac{Iyy_2}{D22^2}-m_1-m_7+\frac{Iyy_1+m_7\cdot D13^2}{(D12+D13)^2})\cdot \frac{L7}{2}$ \\
		\item $I_3''=I_3'+m_2'\cdot\begin{bmatrix}
                   	(\frac{L7}{2})^2	&	0			&	0	\\
						0			&	(\frac{L7}{2})^2	&	0	\\
						0			&	0			&	0
                   \end{bmatrix}$	\\
		\phantom{-}\\
		In particular, the second moments of inertia of body 3 will be:
		\item $Ixx_3''=Ixx_3+(m_1+m_7-\frac{Iyy_1+m_7\cdot D13^2}{(D12+D13)^2}+m_2-\frac{Iyy_2}{D22^2})\cdot (\frac{L7}{2})^2$
		\item $Iyy_3''=Iyy_3+(m_1+m_7-\frac{Iyy_1+m_7\cdot D13^2}{(D12+D13)^2}+m_2-\frac{Iyy_2}{D22^2})\cdot (\frac{L7}{2})^2$
		\item $Izz_3''=Izz_3$
	\end{itemize}

	\item \textbf{Joint 5-3}: Conditions 1, 2 and 5 are satisfied, and a monopole, a dipole and a quadrupole are transferred from body 5 to body 3.
	\begin{itemize}
		\item $m_3'''=m_3''+m_5=m_3+m_1+m_7-\frac{Iyy_1+m_7\cdot D13^2}{(D12+D13)^2}$
		\item $\phantom{m_3'''}+m_2-\frac{Iyy_2}{D22^2}+m_5$ \\
		\phantom{-}\\
		Addition of mass $m_5$ to body 3 at point $\textbf{K}$ also contributes to the first moments of inertia of body 3.\\
		\item $mx_3'''=mx_3''+m_5\cdot DKx=mx_3+m_5\cdot DKx$ \\
		\item $mz_3'''=mz_3''+m_5\cdot (DKz+L7/2)= mz_3+(m_2-\frac{Iyy_2}{D22^2}-m_1-m_7+\frac{Iyy_1+m_7\cdot D13^2}{(D12+D13)^2})\cdot \frac{L7}{2}+m_5\cdot (DKz+L7/2)$ \\
		\phantom{-}\\
		Moreover, the first moment of inertia of body 5 in its $z$ direction ($mz_5$) is also transferred to body 3.
		\item $my_3'''=my_3''+m_5\cdot DKy-mx_5$
		\item $\phantom{my_3'''}=my_3+m_5\cdot DKy-mx_5$ \\
		\phantom{-}\\
		When transferring a monopole, a dipole and a quadrupole from body 5 to 3, the resulting inertia tensor of body 3 will make contributions regarding the three multipoles. The three contributions will be denoted as  $\overline{\overline{m_5}},\overline{\overline{mx_5}}$ and $\overline{\overline{Izz_5}}$, respectively.

		Being $\overline{P_3K}=(DKx,DKy,DKz+\frac{L7}{2})^T$,

% 		\item $\overline{\overline{m_5}}=m_5 \begin{bmatrix}
% 		DKy^2+(DKz+\frac{L7}{2})^2	&	-DKx\cdot DKy			&	-DKx\cdot (DKz+\frac{L7}{2})	 \\
% 		-DKx\cdot DKy			&	DKx^2+(DKz+\frac{L7}{2})^2	&	-DKy\cdot (DKz+\frac{L7}{2})	 \\
% 		-DKx\cdot (DKz+\frac{L7}{2})	&	-DKy\cdot (DKz+\frac{L7}{2})	&	 DKx^2+DKy^2
% 											\end{bmatrix}$

		\item $\overline{\overline{m_5}}=-m_5 \widetilde{\overline{P_3K}} \,\widetilde{\overline{P_3K}}$

		\item $\overline{\overline{mx_5}}=
			mx_5 \begin{bmatrix}
				-2\cdot DKy	&	DKx				&	0				\\
				DKx			&	0				&	DKz+\frac{L7}{2}	\\
				0			&	DKz+\frac{L7}{2}	&	-2\cdot DKy
				\end{bmatrix}$

		\item $\overline{\overline{Izz_5}}=
			Izz_5 \begin{bmatrix}
				1	&	0	&	0	\\
				0	&	0	&	0	\\
				0	&	0	&	1
				\end{bmatrix}$

		\item $I_3'''=I_3''+\overline{\overline{m_5}}+\overline{\overline{mx_5}}+\overline{\overline{Izz_5}}$

		\item $m_5'=0$ \\
		\item $mx_5'=0$ \\
		\item $Iyy_5'=Iyy_5-Izz_5$ \\
		\item $Izz_5'=0$
	\end{itemize}

	\item \textbf{Joint 4-3}: Condition 1 is satisfied and the complete mass of body 4 is transferred to body 3.
	\begin{itemize}
		\item $m_4''=0$ \\
		\item $m_3^{iv}=m_3'''+m_4'=m_4-\frac{Iyy_4}{L10^2}+m_3+m_2-\frac{Iyy_2}{D22^2} +m_5+m_1+m_7-\frac{Iyy_1+m_7\cdot D13^2}{(D12+D13)^2}$ \\
		\item $mx_3^{iv}=mx_3'''+m_4'\cdot D31$
		\item $\phantom{mx_3^{iv}}=mx_3+m_5\cdot DKx+(m_4-\frac{Iyy_4}{L10^2})\cdot D31$ \\
		\item $I_3^{iv}=I_3'''+(m_4-\frac{Iyy_4}{L10^2})\cdot\begin{bmatrix}
		                   	0		&	0		&	0	\\
					0		&	D31^2		&	0	\\
					0		&	0		&	D31^2
		                   \end{bmatrix}$
	\end{itemize}

	\item \textbf{Joint 6-7}: Condition 6 is satisfied and the complete inertia tensor is transferred from body 7 to body 6.
	\begin{itemize}
		\item $I_6'=I_6+I_7$	\\
		\item $I_7'=0$
	\end{itemize}

\end{itemize}
The derived multipole transfers lead to the resulting symbolic base parameters expressions listed in Tab.~\ref{tab::tabla_parametros_base}.

\end{document}